# Post-Editing Error Correction Algorithm For Speech Recognition using Bing Spelling Suggestion

Youssef Bassil, Mohammad Alwani
LACSC – Lebanese Association for Computational Sciences
Beirut, Lebanon

*Abstract*—ASR short for Automatic Speech Recognition is the process of converting a spoken speech into text that can be manipulated by a computer. Although ASR has several applications, it is still erroneous and imprecise especially if used in a harsh surrounding wherein the input speech is of low quality. This paper proposes a post-editing ASR error correction method and algorithm based on Bing's online spelling suggestion. In this approach, the ASR recognized output text is spell-checked using Bing's spelling suggestion technology to detect and correct misrecognized words. More specifically, the proposed algorithm breaks down the ASR output text into several word-tokens that are submitted as search queries to Bing search engine. A returned spelling suggestion implies that a query is misspelled; and thus it is replaced by the suggested correction; otherwise, no correction is performed and the algorithm continues with the next token until all tokens get validated. Experiments carried out on various speeches in different languages indicated a successful decrease in the number of ASR errors and an improvement in the overall error correction rate. Future research can improve upon the proposed algorithm so much so that it can be parallelized to take advantage of multiprocessor computers.

## I. INTRODUCTION

With the ever increasing number of computer-based applications, modern digital computers are no more solely used for crunching numbers and performing high-speed mathematical computations. Instead, they are currently being used for a wider spectrum of tasks including gaming, sound editing, text editing, aided-design, industrial control, medical diagnosis, communication, and information sharing over the World Wide Web. As a matter of fact, computer scientists and researchers from all over the globe have been rigorously carrying out novel innovations and developing groundbreaking solutions to automate every area of life. Automatic Speech Recognition (ASR) is one of the most evolving computing fields that has already been exhaustively employed for an assortment of applications including but not limited to automated telephone services (ATS), voice user interface (VUI), voice-driven industrial control systems (ICS), speech-driven home automation systems (Domotics), speech dictation systems, and automatic speech-to-text systems (STT). Lately, ASR has been of great attraction to computer researchers, manufacturers, and consumers [1]. At heart, the task of ASR is to transform an acoustic waveform into a string of words that can be manipulated by a computing machine [2]. It is thereby abridging the complexity of man-machine interface (MMI) by replacing conventional input devices with an easier, faster, more efficient, and more natural method, allowing users to seamlessly operate, control, and manage computer systems [3]. However, ASR systems are still error-prone and inaccurate especially if they are deployed in an inadequate environment [4]. Generally, ASR errors are manifested by lexical misspellings and linguistic mistakes in the recognized output text, and are primarily caused by the excessive noise in the surroundings, the quality of the speech, the dialect and the utterance of the discourse, and the vocabulary size of the ASR system [4], [5].

In an attempt to reduce the number of errors generated by ASR systems and improve their accuracy, several error-correction techniques were envisioned, some of which are manual as they post-edit the recognized output transcript to correct misspellings; while, others are enhanced acoustic mathematical models aimed at improving the interpretation of the input waveform to prevent errors at early stages [6]. Despite all these endeavors, ASR errors are still at their peak as the mainstream error-correction algorithms are still far from perfect and word errors in speech recognition are always the rule, rather than the exception.

This paper proposes an automatic post-editing context-based real-word error correction approach based on Bing web search engine's spelling suggestion technology [7], to detect and correct linguistic and lexical errors generated by ASR recognition systems. Post-editing (i.e. post-processing) implies that detecting and correcting errors are done after the input wave has been transformed into text. Algorithmically, after the speech has been recognized and converted into text, a list of word-tokens t are generated from the text and then sent successively to Bing search engine as search queries. If Bing returns an alternative spelling suggestion for $t_i$ in the form of "Including results for $c_i$", where $c_i$ is the suggested correction for $t_i$, then $t_i$ is said to contain some misspelled words and $c_i$ is its predictable substitute correction; otherwise no correction is needed for $t_i$ and the next token is processed. Ultimately, when all tokens get validated, all the initial correct tokens { $t_1…t_n$ }, in addition to the corrected ones { $c_1…c_m$ } are concatenated together, yielding to a new transcript with fewer misspellings.

## II. ERRORS IN ASR SYSTEMS

Despite the latest developments in ASR systems, they still exhibit misspellings and linguistic errors in the output text. An evaluation conducted at IBM [8] to measure the number of errors generated by speech recognition dictation systems that were operated by IBM employees, showed that these systems were committing an average of 105 errors per minute, most of which can only be corrected by manually post-editing the text after the end of the speaking. In effect, these errors are caused by two factors: external and internal factors.

*A. External Factors*

The noise in the environment is one of the most key external factors that determine the error rate in ASR systems. If the recognition process is to occur in a quiet location rather than in a noisy open place, superior and accurate results can be attained. It is worth noting that in addition to the raw noise in the setting, the quality of the input devices has a collateral influence in raising the SNR (Signal-to-Noise) ratio. For this reason, high-quality expensive microphones and audio systems are often used to subtly filter the background noise and eliminate the Hiss effect in the input signal which eventually helps exalting the overall precision of the ASR system.

Another weighty factor that needs to be considered is the type of speech being recognized; it is either isolated-word speech or continuous speech. In effect, the recognition of isolated-word speech such as control, telephony, and voice user interface systems is far much easier than the recognition of continuous speech such as dictation or translation systems due to the abundance of pauses in the discourse which makes it less complicated to process and less resource demanding.

Last but not least, the dialect and the speech utterance have a weighty effect on ASR errors. In fact, the dialect of a language varies from epoch to epoch, from country to country, from region to region, and from speaker to speaker. Basically, the accent of non-native speakers makes it harder for ASR systems to recognize and interpret the speech. It has been reported that the error rate is four times higher for non-native speakers than for native speakers [9]. Moreover, the way words are uttered has a direct impact on the recognition system as a whole, for instance, people with a quivering and wavering voice such as children and handicapped may create some hitches during the recognition process.

*B. Internal Factors*

The Internal factors that are responsible for the emergence of ASR errors typically arose from within the components of ASR systems. Inherently, an ASR system is composed of an acoustic model (AM) based on a phonetic lexicon, and a language model (LM) based on an n-gram lexicon [10], [11], [12].

The acoustic model (AM) which computes the likelihood of the observed input phoneme given linguistic units (phones) is based on a lexicon or a dictionary of words with their corresponding phones and pronunciations. These phones are used to recognize the spoken words. Consequently, a deficiency in the dictionary to cover all possible pronunciations would prevent the system from correctly identifying the words in a speech. This situation is often referred to as out-of-vocabulary (OOV) which usually occurs when an ASR system is unable to match a spoken word with any of the entries in its phonetic lexicon [10].

The language model (LM) which approximates how likely a given word is next to occur in a particular text, depends on a probabilistic n-gram model trained on specific corpus of text to predict the next word in a sequence of spoken words. Since it is practically impossible to find a corpus containing all valid words of a language, mismatches and ambiguities would befall during speech recognition, leading subsequently to an increase in the ASR error rate.

As a result and since an ASR system is exclusively based on two types of lexicons; one phonetic of static pronunciations and one probabilistic of n-gram collocations, the larger the vocabulary these lexicons have, the more accurate and the least erroneous the recognition process is considered to be.

### III. RELATED WORK

Different error correction techniques exist, whose purpose is to detect and correct misspelled words generated by ASR systems. Broadly, they can be broken down into several categories: Manual error correction, error correction based on alternative hypothesis, error correction based on pattern learning, and post-editing error correction.

In manual error correction, a staff of people is hired to review the output transcript generated by the ASR system and correct the misspelled words manually by hand. This is to some extent considered laborious, time consuming, and error-prone as the human eye may miss some errors.

Another category is the alternative hypothesis error correction in which an error is replaced by an alternative word-correction called hypothesis. The chief drawback of this method is that the hypothesis is usually derived from a lexicon of words; and hence it is susceptible to high out-of-vocabulary rate. In that context, Setlur, Sukkar, and Jacob [13] proposed an algorithm that treats each utterance of the spoken word as hypothesis and assigns it a confidence score during the recognition process. The hypothesis that bypasses a specific threshold is to be selected as the correct output word. The experiments showed that the error rate was reduced by a factor of 0.13%.

Likewise, Zhou, Meng, and Lo [14] proposed another algorithm to detect and correct misspellings in ASR systems. In this approach, twenty alternative words are generated for every single word and treated as utterance hypotheses. Then, a linear scoring system is used to score every utterance with certain mutual information, calculated from a training corpus. This score represents the number of occurrence of this specific utterance in the corpus. After that, utterances are ranked according to their scores; the one with the highest score is chosen to substitute the detected error. Experiments conducted, indicated a decrease in the error rate by a factor of 0.8%.

Pattern learning error correction is yet another type of error correction techniques in which error detection is done through finding patterns that are considered erroneous. The system is first trained using a set of error words belonging to a specific domain. Subsequently, the system builds up detection rules that can pinpoint errors once they occur. At recognition time, the ASR system can detect linguistic errors by validating the output text against its predefined rules.

(IJACSA) International Journal of Advanced Computer Science and Applications,
Vol. 3, No.2, 2012The drawback of this approach is that it is domain specific; and thus, the number of words that can be recognized by the system is minimal. In this perspective, Mangu and Padmanabhan [15] proposed a transformation-based learning algorithm for ASR error correction. The algorithm exploits confusion network to learn error patterns while the system is offline. At run-time, these learned rules assist in selecting an alternative correction to replace the detected error. Similarly, Kaki, Sumita, and Iida, [16] proposed an error correction algorithm based on pattern learning to detect misspellings and on similarity string algorithm to correct misspellings. In this technique, the output recognized transcript is searched for potential misspelled words. Once an error pattern is detected, the similarity string algorithm is applied to suggest a correction for the error word. Experiments were executed on a Japanese speech and the results indicated an overall 8.5% reduction in ASR errors. In a parallel effort, statistical-based pattern learning techniques were also developed. Jung, Jeong, and Lee [17] employed the noisy channel model to detect error patterns in the output text. Unlike other pattern learning techniques which exploit word tokens, this approach applies pattern learning on smaller units, namely individual characters. The global outcome was a 40% improvement in the error correction rate. Furthermore, Sarma and Palmer [18] proposed a method for detecting errors based on statistical co-occurrence of words in the output transcript. The idea revolves around contextual information which states that a word usually appears in a text with some highly co-occurred words. As a result, if an error occurs within a specific set of words, the correction can be statistically deduced from the co-occurred words that often appear in the same set.

The final type of error correction is post-editing. In this approach, an extra layer is appended to the ASR system with the intention of detecting and correcting misspellings in the final output text after recognition of the speech is completed. The advantage of this technique is that it is loosely coupled with the inner signal and recognition algorithms of the ASR system; and thus, it is easy to be implemented and integrated into an existing ASR system while taking advantage of other error correction explorations done in sister fields such as OCR, NLP, and machine translation.

As an initial attempt, Ringger and Allen [19] proposed a post-processor model for discovering statistical error patterns and correct errors. The post-processor was trained on data from a specific domain to spell-check articles belonging to the same domain. The actual design is composed of a channel model to detect errors generated during the speech recognition phase, and a language model to provide spelling suggestions for those detected errors. As outcome, around 20% improvement in the error correction rate was achieved. On the other hand, Ringger and Allen [20] proposed a post-editing model named SPEECHPP to correct word errors generated by ASR systems. The model uses a noisy channel to detect and correct errors, in addition to the Viterbi search algorithm to implement the language model. Another attempt was presented by Brandow and Strzalkowski [21] in which, text generated from the ASR system is collected and aligned with the correct transcription of the same text. In a training process, a set of correction rules are generated from these transcription texts and validated against a generic corpus; rules that are void or invalid are discarded. The system loops for several iterations until all rules get verified. Finally, a post-editing stage is employed which harnesses these rules to detect and correct misspelled words in the ASR generated transcript.

IV. PROPOSED SOLUTION

This paper proposes a new post-editing approach and algorithm for ASR error correction based on Bing's spelling suggestion technology [7]. The idea hinges around using Bing's enormous indexed data to detect and correct real-word errors that appear in the ASR recognized output text. In other words, error correction is applied to spell check the final text that resulted from the transformation of the input wave into text; and hence is referred to as post-editing error correction. The algorithm starts first by chopping the ASR output text into several word tokens. Then, each token is sent to Bing's web search engine as a search query. If this query contains a misspelled word, Bing suggests a spelling correction for it, and consequently, the algorithm replaces it with the spelling suggestion.

A. Bing's Spelling Suggestion

Bing's spelling suggestion technology can suggest alternative corrections for the often made typos, misspellings, and keyboarding errors. At the core, Bing has a colossal database of billions of online web pages containing trillions of term collections and n-gram words that can be used as groundwork for all kinds of linguistic applications such as machine translation, speech recognition, spell checking, as well as other types of text processing problems. Fundamentally, Bing's spelling suggestion algorithm is based on the probabilistic n-gram model originally proposed by Markov [22] for predicting the next word in a particular sequence of words. In brief, an n-gram is simply a collocation of words that is n words long.

For instance, "The boy" is a 2-gram phrase also referred to as bigram, "The boy scout" is a 3-gram phrase also referred to as trigram, "The boy is driving his car" is a 6-gram phrase, and so forth. The Bing's algorithm automatically examines every single word in the search query for any possible misspellings. It tries first to match the query, basically composed of ordered association of words, with any occurrence alike in Bing's database of indexed web pages; if the number of occurrence is high, then the query is considered correct and no correction is to take place.

However, if the query was not found, Bing uses its n-gram statistics to deduce the next possible correct word in the query. Sooner or later, an entire suggestion for the whole misspelled query is generated and displayed to the user in the form of "Including results for spelling-suggestion". For example, searching for the word "conputer" drives Bing to suggest "Including results for Computer". Likewise, searching for "The hord disk sturage" drives Bing to suggest "Including results for the hard-disk storage". Searching for the proper name "jahn cenedy" drives Bing to suggest "Including results for John Kennedy". Figure 1-3 show the spelling suggestions returned by Bing search engine when searching for "conputer", "The hord disk sturage", and "jahn cenedy" respectively.

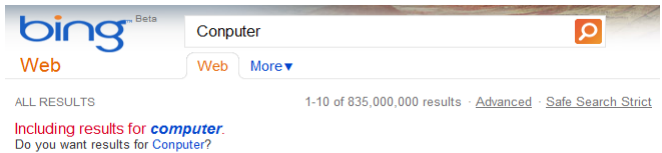

Figure 1. Spelling suggestion for "conputer"

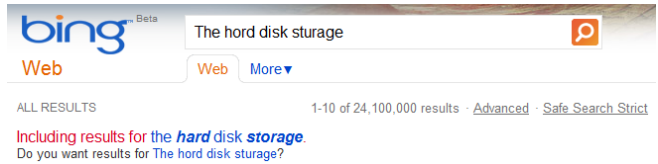

Figure 2. Spelling suggestion for "The hord disk sturage"

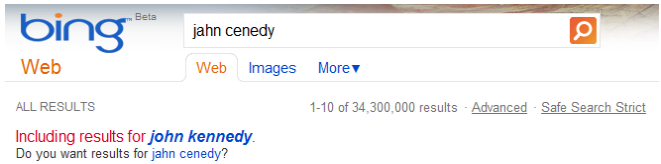

Figure 3. Spelling suggestion for "jahn cenedy"

*B. ASR Model*

The proposed error correction method is executed during the post-editing stage of an ASR system, and is based on Bing's spelling suggestion. At early stages, prior to post-editing the recognized text using the proposed algorithm, a standard ASR system is fed by an input waveform that represents the speech to be recognized. Then, the signal is digitally processed in order to extract its spectral features and audible phones. Afterwards, the likelihood of an observed phoneme given an extracted spectral feature is computed by the acoustic model (AM) and its Hidden Markov Model (HMM) and phonetic lexicon. In parallel, the language model (LM) computes the probability of the obtained phone to occur in the language. Finally, a decoding module statistically infers the spoken words and generates the final output text.

The proposed model further processes the obtained output and adds a post-editing stage to the system with the purpose of detecting and correcting any possible misspelled words that were generated during the recognition process. In essence, the output text that is obtained from the decoding module is broken down into a collection of tokens, each made out of six words. In a sequential fashion, these tokens are sent one after the other to Bing search engine as search parameters. If Bing does not return a spelling suggestion, then it is evident that the query contains no misspelled words; and thus no correction is needed for this particular token and no changes is to occur for the original text.

On the other hand, if Bing returns a spelling suggestion, then definitely the query contains some misspelled words; and thus a correction is required for this particular token of words. The correction consists of replacing the original token in the text by the Bing's suggested correction. Figure 4 depicts a block diagram for a generic ASR system, however modified by adding to it a post-editing layer to perform error correction using Bing's spelling suggestion.

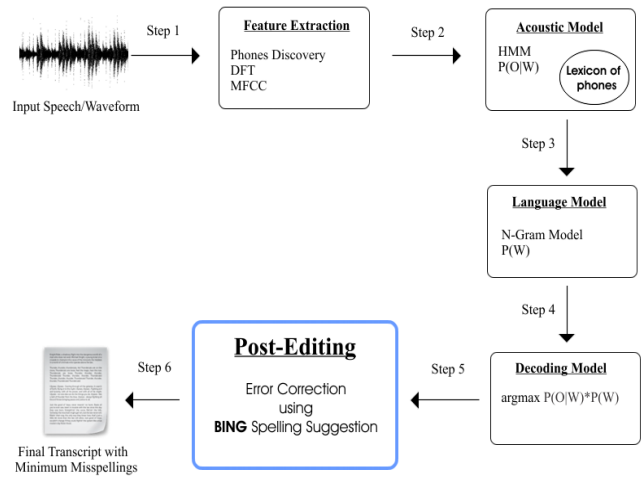

Figure 4. ASR system with a post-editing layer for error correction

*C. The Error Correction Algorithm*

The proposed algorithm comprises several steps to be executed in order to detect and correct *ASR* misspellings. The algorithm takeoffs by dividing the recognized output transcript into several tokens $T=\{t_1…t_n\}$, each composed of 6 words, $t_i=\{w_0,w_1,w_2,w_3,w_4,w_5\}$ where $t_i$ is a particular token and $w_j$ is a single word or term in that token. Then, every $t_i$ is sent to be validated using Bing search engine. The search results returned by Bing are then parsed to identify whether or not they contain the "*Including results for $c_i$*" spelling suggestion message, where $c_i$ is the suggested correction for $t_i$. If true, then token $t_i$ must contain a certain misspelled word; and hence, $t_i$ is replaced by $c_i$. Ultimately, after all tokens get validated, all original correct tokens $O=\{t_1…t_k\}$, plus the corrected ones $C=\{c_1…c_p\}$ are concatenated together, yielding to a new text with fewer misspellings represented formally as $V=\{v_1…v_{k+p}\}$. The post-editing process then finishes and the algorithm halts. Figure 5 summarizes the flow of execution for the proposed algorithm.

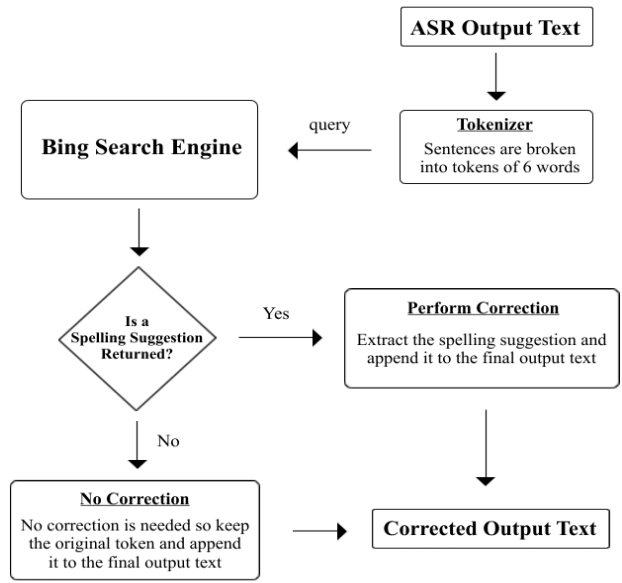

Figure 5. Flowchart illustrating the different steps of the proposed algorithm

*D. The Pseudo-Code*

The following pseudo-code describes the entire logic behind the proposed algorithm, independently of any programming language platform.

```
// the purpose of this procedure is to correct ASR spelling
errors using Bing spelling suggestion
// INPUT: ASR recognized text possibly containing errors and
misspellings
// OUTPUT: Corrected text
START
Procedure Post-Editing(asr_text)
{
    // breaks the asr_text into blocks of 6 words each
    tokens ← Tokenize(asr_text, 6)

    // iterates until all tokens are exhausted
    for (i←0 to tokens_length)
    {
        // send tokens[i] to Bing search engine
        results ← BingSearch(tokens[i])
                    if(results contains("Including results
        for")
            // indicates some misspellings in tokens[i]
            output ← getSuggestion(results)
            // extract correction and append it to output file
        else
            output ← tokens[i]
            // no misspellings so add the original tokens[i]
    }
    RETURN output
}
FINISH
```

The procedure *Post-Editing()* contains one for loop that is executed *n* times, where *n* is the total number of tokens in the *ASR* text. Considering "*results ← BingSearch(tokens[i])*" as the basic operation, the time complexity of the algorithm is described as follows:

$$\sum_{i=0}^{n} 1 = n \text{ and thus the algorithm is of time complexity } O(n)$$

Since the basic operation is to be executed *n* times regardless of the content of the input *ASR* text, $C_{Best}(n)= C_{Worst}(n)= C_{Average}(n)= n = $ number of tokens in the original *ASR text*

## V. EXPERIMENTS & RESULTS

In the experiments, speech recognition was performed on two speeches in two different languages: English [23] and French [24]. Bing.com was used to post-edit the English speech, while Bing.fr was used to post-edit the French speech. As for the ASR software, a custom proprietary application program based on Microsoft Speech Application Programming Interface (SAPI 5.0) engine [25] was utilized to perform the speech recognition of the two input speeches.

The proposed post-editing algorithm was implemented using MS C# 4.0 under the MS .NET Framework 4.0 and the MS Visual Studio 2010.

The following paragraph is the input English speech to be processed by the ASR software.

Virtual machine applications such as VMWare Workstation, Sun Virtualbox, and Microsoft Virtual PC now allow you to boot the second operating system on top of your main OS, eliminating the need and hassle of rebooting into another OS. Installing a NICs driver into a Windows, Macintosh, or Linux system is easy: just insert the driver CD when prompted by the system. Unless you have a very offbeat NIC, the operating system will probably already have the driver preinstalled, but there are benefits to using the driver on the manufacturer CD. IEEE could use the traditional Physical layer mechanisms defined by the Ethernet standard. But, there was already in place a perfectly usable 10 Gbps fiber network, called SONET, used for wide area networking (WAN) transmissions. Microsoft pushed the idea of a single client tunneling into a private LAN using software. Cisco, being the router king that it is, came up with its own VPN protocol called Layer 2 Tunneling Protocol

The subsequent paragraph represents the output transcript generated by the ASR system along with all the misspellings (underlined) that were produced during the recognition process.

Virtual machine applications such as VWare Workstation, Sun Virualbox, and Micro soft Virsual PC now allow you to boot the second operaing system on tat of your main OS, eliminating the need and hassl of reboting into another OS. Installing a NICs driver into a Windoos, Makintosh, or Linix system is easy: just insert the driver CD when promptd by the system. Unless you have a very offbeet NIC, the operating system will probably already have the driver pre-installed, but there are benefits to using the driver on the manufachurer CD. IEEE could use the traditional Physical layer mechanisms defined by the Ethernit standard. But, there was already in place a perfectly usable 10 Gbps fiber network, called SONETT, used for wide area networking (WAAN) transmissions. Micro soft pushed the idea of a single client tullleing into private LAAN using software. Ciskow, being the router king that it is, came up with its own VPN protocol called Layer 2 Tunneling Protocol

Next is the same previous transcript, however, error-corrected using the proposed post-editing error correction algorithm. Underlined are the words that were not corrected.

Virtual machine applications such as VMWare Workstation, Sun Virtualbox, and Microsoft Virtual PC now allow you to boot the second operating system on tat of your main OS, eliminating the need and hassl of rebooting into another OS. Installing a NIC driver into a Windows, Macintosh, or Linux system is easy: just insert the driver CD when prompts by the system. Unless you have a very offbeat NIC, the operating system will probably already have the driver preinstalled, but there are benefits to using the driver on the manufacturer CD. IEEE could use the traditional Physical layer mechanisms defined by the Ethernet standard. But, there was already in place a perfectly usable 10-Gbps fiber network, called SONETT, used for wide area networking (WAN) transmissions. Microsoft pushed the idea of a single client tulle long into a private LAN using software. Cisco, being the router king that it is, came up with its own VPN protocol called Layer 2 Tunneling Protocol

The English recognized transcript comprehended 23 misspelled words out of 161 total words (number of words in the whole speech), making the error rate close to E = 23/161 = 0.142 = 14.2%. Several of these errors were proper names such as "Microsoft", others were technical words such as "LAN", "Macintosh", "Linux", "VMWare", and "Ethernet", and the remaining ones were regular English words such as "virtual", "operating", "rebooting", "hassle", etc. When the proposed post-editing error correction algorithm was applied, 18 misspelled words out of 23 were corrected successfully, leaving only 5 non-corrected errors and they were as follows: "promptd" was miss-corrected as "prompts", "tulleing" was miss-corrected as "tuelle long", and "tat", "hassl", and "SONETT" were not corrected at all. As a result, the error rate using the proposed algorithm was close to E = 5/161 = 0.031 = 3.1%. Consequently, the improvement can be calculated as I = 0.142/0.031 = 4.58 = 458%, that is increasing the rate of error detection and correction by a factor of 4.58.

Another experiment was conducted on a French speech and it is delineated below:

Enfin pour nuancer les sens attribué à ces quatre directions de l'espace, M. Monod a proposé une combinaison du haut et du bas avec l'orientation à droite et à gauche. Les deux zones gauches sont charactérisées par des élément plus passifs dans la psychologie de l'individu et sont associées au passé de celui ci dans la zone bas gauche. Dans la zone droite on retrouve la même distinction entre les facteurs les plus dynamiques en haut droit et les processus de socialisation plus anciens en bas droit. Les crayons de couleur constituent un stimulus banal d'ou son impact sur le sujet est moins fort que celui des planches d'encre de Rorschach.

The subsequent paragraph represents the output transcript generated by the ASR system along with all the misspellings (underlined) that were produced during the recognition process.

Enfin pur nuancer les sence attribué à ces quatre directions de l'espace, M. Mono a proposé une combinaison du haut et du bas avec l'orientation à droite et à gouche. Les deux zones gouches sont charactérisées par des élément plus passivs dans la psycologie de l'indivitu et sont associées au passé de celui ci dans la zone bas gouche. Dans la zone droite on retruve la même distinction entre les facdeurs les plus dynamiques en haot droit et les processuse de socialisation plus anciens en bas droit. Les craiyons de couleur constituent un stimulus banal dou son impact sur le sujet est moins fort que celui des planches d'encre de Roschah.

Next is the same previous transcript, however error-corrected using the proposed post-editing error correction algorithm. Underlined are the words that were not corrected.

Enfin pour nuancer les sens attribué à ces quatre directions de l'espace, M. Mono a proposé une combinaison du haut et du bas avec l'orientation à droite et à gauche. Les deux zones gauches sont charactérisées par des élément plus passive dans la psychologie de l'individu et sont associées au passé de celui ci dans la zone bas gauche. Dans la zone droite on retrouve la même distinction entre les facteurs les plus dynamiques en haut droit et les processus de socialisation plus anciens en bas droit. Les crayons de couleur constituent un stimulus banal dou son impact sur le sujet est moins fort que celui des planches d'encre de Rorschach.

The French recognized transcript comprehended 16 misspelled words out of 110 total words (number of words in the whole speech), making the error rate close to E = 16/110 = 0.145 = 14.5%. Several of these errors were proper names such as "Rorschach", and others were regular French words such as "pour", "gauche", "retrouve", "crayons", etc. When the proposed post-editing error correction algorithm was applied, 13 misspelled words out of 16 were corrected successfully, leaving only 3 non-corrected errors and they were as follows: "passivs" was miss-corrected as "passive", and "Mono" and "dou" were not corrected at all. As a result, the error rate using the proposed algorithm was close to E = 3/110 = 0.027 = 2.7%. Consequently, the improvement can be calculated as I = 0.145/0.027 = 5.37 = 537%, that is increasing the rate of error detection and correction by a factor of 5.37.

VI. EXPERIMENTS EVALUATION

The experiments conducted on the proposed ASR post-editing error correction algorithm evidently showed a 458% improvement in the error correction rate for English speech and 537% for French speech. In other terms, around 4.5 times more English errors were detected and corrected, and 5.3 times more French errors were detected and corrected. On average, the proposed algorithm improved the error correction rate by I = (458% + 537%) / 2 = 497%, that is increasing the overall rate of error detection and correction by a factor of 4.9. Table 1 summarizes the experimental results obtained for the proposed ASR error correction algorithm before and after post-editing.

TABLE I. EXPERIMENTAL RESULTS BEFORE AND AFTER POST-EDITING

|  | **English Document** **Total words = 161** | **French Document** **Total words = 110** |
|---|---|---|
| Number of errors resulted before post-editing | 23 | 16 |
| Number of errors resulted after post-editing | 5 | 3 |
| Error rate before post-editing | 14.2% | 14.5% |
| Error rate after post-editing | 3.1% | 2.7% |
| Improvement ratio | 4.58 (458%) | 5.37 (537%) |

VII. CONCLUSIONS

This paper presented a new ASR post-editing error correction method based on Bing's online spelling suggestion technology. The backbone of this technology is a large dataset of words and sentences indexed by Bing and originally extracted from several online sources including web pages, documents, articles, and forums. This allows Bing to suggest common spellings for queries containing errors and linguistic mistakes. For this reason, the proposed algorithm excelled in detecting and correcting ASR errors as it fully harnessed Bing's online spelling suggestion to spell-check the ASR recognized output text. Experiments carried out, indicated a noticeable reduction in the number of ASR errors, yielding to an outstanding improvement in the ASR error correction rate.

## VIII. FUTURE WORK

As future work, various ways to parallelize the proposed algorithm are to be investigated so as to take advantage of multiprocessors and distributed computers. The projected outcome would be a faster algorithm of time complexity $O(n/p)$, where n is the total number of word tokens to be spell checked and p is the total number of processors.

ACKNOWLEDGMENT

This research was funded by the Lebanese Association for Computational Sciences (LACSC), Beirut, Lebanon under the "Web-Scale Speech Recognition Research Project – WSSRRP2011".

REFERENCES

[1] L. Rabiner and B.H. Juang, Fundamentals of Speech Recognition, Prentice Hall PTR, Englewood Cliffs, New Jersey, 1993.
[2] T. Dutoit, An Introduction to Text-to-Speech Synthesis (Text, Speech and Language Technology), Springer, 1st ed, 1997.
[3] W. A. Lea, "The value of speech recognition systems", Readings in Speech Recognition, Morgan Kaufmann Publishers, Inc, San Francisco, CA, pp. 39–46, 1990.
[4] L. Deng and X. Huang, "Challenges in adopting speech recognition", Communications of the ACM, vol. 47, No.1, pp.60–75, 2004.
[5] D. Jurafsky, J. Martin, Speech and Language Processing, 2nd ed, Prentice Hall, 2008.
[6] A. Rudnicky, A. Hauptmann, and K. Lee, "Survey of Current Speech Technology", Communications of the ACM, vol. 37, No. 3, pp.52-57, 1994.
[7] Microsoft Corporation, Microsoft Bing Search Engine, URL: http://www.bing.com, 2011.
[8] J.R. Lewis, "Effect of Error Correction Strategy on Speech Dictation Throughput", Proceedings of the Human Factors and Ergonomics Society, pp.457-461, 1999.
[9] Tomokiyo, L. M, "Recognizing non-native speech: Characterizing and adapting to non-native usage in speech recognition", Ph.D. thesis, Carnegie Mellon University, 2001.
[10] L.L. Chase, "Error-Responsive Feedback Mechanisms for Speech Recognizers", PhD thesis, Robotics Institute, Carnegie Mellon University, Pittsburgh, PA, 1997.
[11] D. Gibbon, R. Moore, and R. Winski, "Handbook of Standards and Resources for Spoken Language Systems", Mouton de Gruyter, 1997.
[12] B. Suhm, "Multimodal Interactive Error Recovery for Non-Conversational Speech User Interfaces", PhD thesis, University of Karlsruhe, 1998.
[13] A. R. Setlur, R. A. Sukkar, and J. Jacob, "Correcting recognition errors via discriminative utterance verification", In Proceedings of the International Conference on Spoken Language Processing, pp.602–605, Philadelphia, PA, 1996.
[14] Z. Zhou, H. M. Meng, and W. K. Lo, "A multi-pass error detection and correction framework for Mandarin LVCSR", In Proceedings of the International Conference on Spoken Language Processing, pages 1646–1649, Pittsburgh, PA, 2006.
[15] L. Mangu and M. Padmanabhan, "Error corrective mechanisms for speech recognition", In Proceedings of the IEEE International Conference on Acoustics, Speech, and Signal Processing, vol. 1, pp.29–32, Salt Lake City, UT, 2001.
[16] S. Kaki, E. Sumita, and H. Iida, "A method for correcting errors in speech recognition using the statistical features of character co-occurrence", In COLING-ACL, pp.653–657, Montreal, Quebec, Canada, 1998.
[17] S. Jung, M. Jeong, and G. G. Lee, "Speech recognition error correction using maximum entropy language model", In Proceedings of the International Conference on Spoken Language Processing, pp.2137–2140, Jeju Island, Korea, 2004.
[18] A. Sarma and D. D. Palmer, "Context-based speech recognition error detection and correction", In Proceedings of the Human Language Technology Conference of the North American Chapter of the Association for Computational Linguistics, pp.85–88, Boston, MA, 2004.
[19] E. K. Ringger and J. F. Allen, "Error correction via a post-processor for continuous speech recognition", In Proceedings of the IEEE International Conference on Acoustics, Speech, and Signal Processing, vol.1, pp.427–430, Atlanta, GA, 1996.
[20] Eric K. Ringger and James F. Allen, "A Fertility Channel Model for Post-Correction of Continuous Speech Recognition", In Proceedings of the Fourth International Conference on Spoken Language Processing, vol.2, pp. 897-900, 1996.
[21] R. L. Brandow and T. Strzalkowski, "Improving speech recognition through text-based linguistic post-processing", United States Patent 6064957, 2000.
[22] A.A. Markov, "Essai d'une Recherche Statistique Sur le Texte du Roman 'Eugène Oneguine'", Bull. Acad. Imper. Sci. St. Petersburg, vol. 7, 1913.
[23] M. Meyers, CompTIA Network+ All-in-One Exam Guide, 4th ed, McGraw-Hill Osborne Media, 2009.
[24] N. Kim-Chi, La Personnalité et l'épreuve de dessins multiples, PUF edition, 1989.
[25] Microsoft Corporation, Microsoft Speech Application Programming Interface (SAPI 5.0) engine, URL: http://www.microsoft.com/download/en/details.aspx?displaylang=en&id=10121, 2011.